%% file: main.tex
\documentclass[10pt,twocolumn,letterpaper]{article}

\usepackage{cvpr}              

\usepackage{algorithm}
\usepackage{algorithmic}
\usepackage{amssymb}
\usepackage{multirow}
\usepackage{booktabs}
\usepackage{xcolor}
\usepackage{amsmath}
\usepackage{pifont}
\usepackage{tabularx}
\usepackage[table]{xcolor}
\usepackage{colortbl}

\input{preamble}
\definecolor{cvprblue}{rgb}{0.21,0.49,0.74}
\usepackage[pagebackref,breaklinks,colorlinks,allcolors=cvprblue]{hyperref}

\def\paperID{}
\def\confName{CVPR}
\def\confYear{2026}

\title{Decompose and Transfer: CoT-Prompting Enhanced Alignment \\for Open-Vocabulary Temporal Action Detection}

\author{Sa Zhu$^{1,2,3}$ \quad Wanqian Zhang$^{1}$ \thanks{Corresponding author} \quad Lin Wang$^{4}$  \quad Xiaohua Chen$^{5}$ \\ \quad  Chenxu Cui$^{1,2,3}$  \quad  Jinchao Zhang$^{1,3}$ \quad Bo Li$^{1,3}$
\\ Institute of Information Engineering,
Chinese Academy of Sciences$^1$ \\ School of Cyber Security, University of Chinese Academy of Sciences
$^2$ \\  State Key Laboratory of Cyberspace Security Defense$^{3}$ \\  Hangzhou Dianzi University$^4$ \quad  Department of Automation, Tsinghua University$^5$
\\ \{zhusa, zhangwanqian, cuichenxu, zhangjinchao, libo\}@iie.ac.cn 
}

\begin{document}
\maketitle
\input{sec/0_abstract}    
\input{sec/1_intro}
\input{sec/2_Related_work}

\input{sec/3_methodology}
\input{sec/4_experiment}

\input{sec/X_suppl}

{
\small
     \bibliographystyle{ieeenat_fullname}
     \bibliography{main}
}


\end{document}

%% file: sec/0_abstract.tex
\begin{abstract}
Open-Vocabulary Temporal Action Detection (OV-TAD) aims to classify and localize action segments in untrimmed videos for unseen categories.
Previous methods rely solely on global alignment between label-level semantics and visual features, which is insufficient to transfer temporal consistent visual knowledge from seen to unseen classes. 
To address this, we propose a Phase-wise Decomposition and Alignment (PDA) framework, which enables fine-grained action pattern learning for effective prior knowledge transfer.  
Specifically, we first introduce the CoT-Prompting Semantic Decomposition (CSD) module, which leverages the chain-of-thought (CoT) reasoning ability of large language models to automatically decompose action labels into coherent phase-level descriptions, emulating human cognitive processes.  
Then, Text-infused Foreground Filtering (TIF) module is introduced to adaptively filter action-relevant segments for each phase leveraging phase-wise semantic cues, producing semantically aligned visual representations.
Furthermore, we propose the Adaptive Phase-wise Alignment (APA) module to perform phase-level visual–textual matching, and adaptively aggregates alignment results across phases for final prediction.
This adaptive phase-wise alignment facilitates the capture of transferable action patterns and significantly enhances generalization to unseen actions. 
Extensive experiments on two OV-TAD benchmarks demonstrated the superiority of the proposed method.
\end{abstract}

%% file: sec/1_intro.tex
\begin{figure}[!t]
  \centering
  \setlength{\abovecaptionskip}{-0.1em}
  \includegraphics[width=1.0\linewidth]{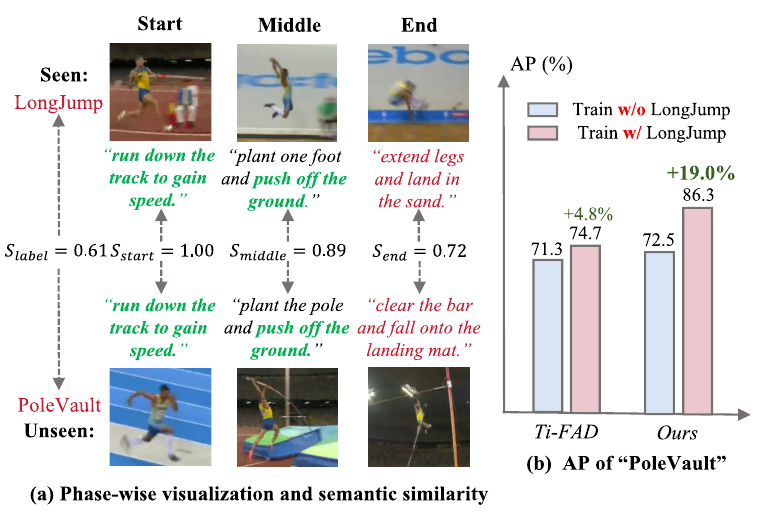}
  \caption{
  Illustrative examples of seen class LongJump and unseen class PoleVault. 
  (a) The “Start” and “Middle” phases of seen LongJump and unseen PoleVault share strong semantic similarities, despite low label similarity. 
  (b) In light of phase-wise prior knowledge, our method shows higher performance on PoleVault when trained with LongJump. 
  }
  \label{Motivation}
  \vspace{-1em}
\end{figure}

\section{Introduction}
\label{sec:intro}
Temporal action detection (TAD)~\cite{zhao2021video, liu2022empirical, zhao2022tuber, he2022asm, tang2023ddg, kim2024te, liu2024end, zhu2024dual, chen2025temporal, pang2025context} aims to classify and localize action instances within untrimmed videos, serving as a fundamental task in video understanding~\cite{hao2025dada++, hao2023dual, fang2025viss, fang2023uatvr, zhu2025uneven, zhu2025endogenous}. 
Conventional TAD approaches~\cite{zhao2017temporal, qing2021temporal, zhang2024hr, kumar2025stable} typically depend on large-scale annotated datasets for supervised training, which is both labor-intensive and time-demanding, limiting their application in practical scenarios. 

To overcome this limitation, Open-Vocabulary Temporal Action Detection (OV-TAD) has emerged, also called as Zero-Shot Temporal Action Detection (ZSTAD)~\cite{nag2022zero, lee2024text, wangconcept}, aiming to detect action instances from categories that are not encountered during training. 
The core challenge lies in establishing a transferable and semantically meaningful association between novel actions and prior semantic knowledge learned from seen classes~\cite{li2024detal}. 
Existing OV-TAD approaches typically construct a shared visual-textual representation space, enabling the model to retrieve the most similar labels based on feature alignment.  
This is commonly facilitated by pre-trained vision-language models (VLMs), such as CLIP~\cite{radford2021learning}, which are trained on large-scale image-text pairs and exhibit strong cross-modal alignment.
Then, they perform action detection through global alignment between label embeddings and visual representations. 
The label with the highest similarity score is then selected as the predicted action category.

Although prior methods have achieved promising performance, globally aligning label-level semantics with visual representations remains inadequate, especially when transferring consistent visual knowledge from seen to unseen classes. 
As illustrated in Figure~\ref{Motivation} (a), actions that are semantically distinct (e.g., LongJump vs. PoleVault) may still share visually similar snippets at fine-grained phase-level.
For instance, the start phase of both actions is `run down the track to gain speed', and the middle phase both entails `pushing off the ground'.
Capturing and representing such shared local action patterns is crucial for knowledge transfer, as these fine-grained visual cues can serve as transferable priors for recognizing novel categories. 
In other words, if the model has been trained on the action LongJump, it is more likely to exhibit improved performance when encountering the unseen action PoleVault, as shown in Figure~\ref{Motivation} (b).
Thus, how to transfer prior visual knowledge from seen to unseen actions becomes the motivation of this work.

In this paper, we propose the Phase-wise Decomposition and Alignment (PDA) framework for OV-TAD, which enables fine-grained action pattern learning for effective knowledge transfer. 
We begin by decomposing action labels using the Chain-of-Thought (CoT) reasoning capability of large language models (LLMs)~\cite{brown2020language, achiam2023gpt}. 
Specifically, we introduce the CoT-Prompting Semantic Decomposition (CSD) module, which emulates human cognition by unfolding actions step by step to generate coherent phase-level descriptions.
Unlike encoding coarse label-level semantics, CSD could capture transferable knowledge among semantically different labels, thereby enhancing generalization to unseen actions.
Besides, to associate textual and visual representations, a naive solution is to apply global alignment by concatenating all phase descriptions and averaging visual features. 
However, this often fails to model fine-grained visual and textual cues that could be transferred for unseen action detection.
To overcome this, we introduce the Text-infused Foreground Filtering (TIF) module, which leverages phase-wise semantic cues to adaptively filter action-relevant segments for each phase, producing semantically aligned visual representations.
Furthermore, we propose the Adaptive Phase-wise Alignment (APA) module to perform phase-wise matching between visual and textual representations and adaptively integrates the alignment results for final prediction.
This phase-wise alignment facilitates the capture of transferable action patterns and significantly enhances generalization to unseen categories.

To summarize, our contributions are as follows:
\begin{itemize}
  \item We propose Phase-wise Decomposition and Alignment (PDA) framework, enabling phase-wise action pattern learning, and facilitating the transferable visual cues for improved generalization in OV-TAD.

  \item We devise the CoT-Prompting Semantic Decomposition (CSD) module, which employs LLMs' CoT capability to automatically decompose action labels into multi-phase descriptions, producing more transferable semantics.  
  
  \item  We further introduce the Text-infused Foreground Filtering (TIF) and the Adaptive Phase-wise Alignment (APA) modules, to perform phase-level cross-modal matching and adaptive integration, enhancing the model’s ability to capture transferable action patterns. 
\end{itemize}


%% file: sec/2_Related_work.tex
\section{Related Work}
\label{sec:related work}


\textbf{Open-Vocabulary Temporal Action Detection} aims to localize and recognize action segments from unseen categories by transferring knowledge from seen actions~\cite{zhang2022tn}.
Efficient-Prompt~\cite{ju2022prompting} introduces activity proposals classified via cosine similarity between proposal features and CLIP-generated label embeddings. 
However, its two-stage design suffers from interference between localization and classification. 
To address this, STALE~\cite{nag2022zero} mitigates error propagation between the stages, while DeTAL~\cite{li2024detal} decouples the tasks entirely using separate networks for proposal generation and classification. 
More recently, Ti-FAD~\cite{lee2024text} introduces a cross-attention mechanism to infuse textual information into visual features, enhancing subsequent classification and regression. 
In parallel, CSP~\cite{wangconcept} projects video features into a semantic concept space to enhance the semantic consistency of learned action representations. 
However, they merely align label-level semantics with global proposal features, ignoring fine-grained temporal knowledge that can be transferred from seen to unseen actions.

\noindent
\textbf{LLM for Label Expansion} on action localization and recognition has been proposed to generate detailed action descriptions, enriching textual semantics and narrowing the modality gap for cross-modal alignment. 
For instance, ~\cite{ju2023multi} decompose actions into defining attributes and match their combinations with frame embeddings for action localization. 
Similarly,  ~\cite{bosetti2024text, jia2024generating} generate multi-dimensional textual descriptions and compute their similarity with averaged visual embeddings for action recognition.
In contrast, we utilize the Chain-of-Thought (CoT) reasoning capability of LLMs to decompose each action into multiple phases, aiming to extract transferable knowledge across semantically diverse labels. 
Combined with adaptive phase-wise alignment, our model learns phase-level action patterns that could be generalized from seen to unseen categories, rather than enriching textual semantics alone.

\begin{figure*}[!t]
\centering
  \setlength{\abovecaptionskip}{-0.1em}
  \includegraphics[width=0.97\textwidth]{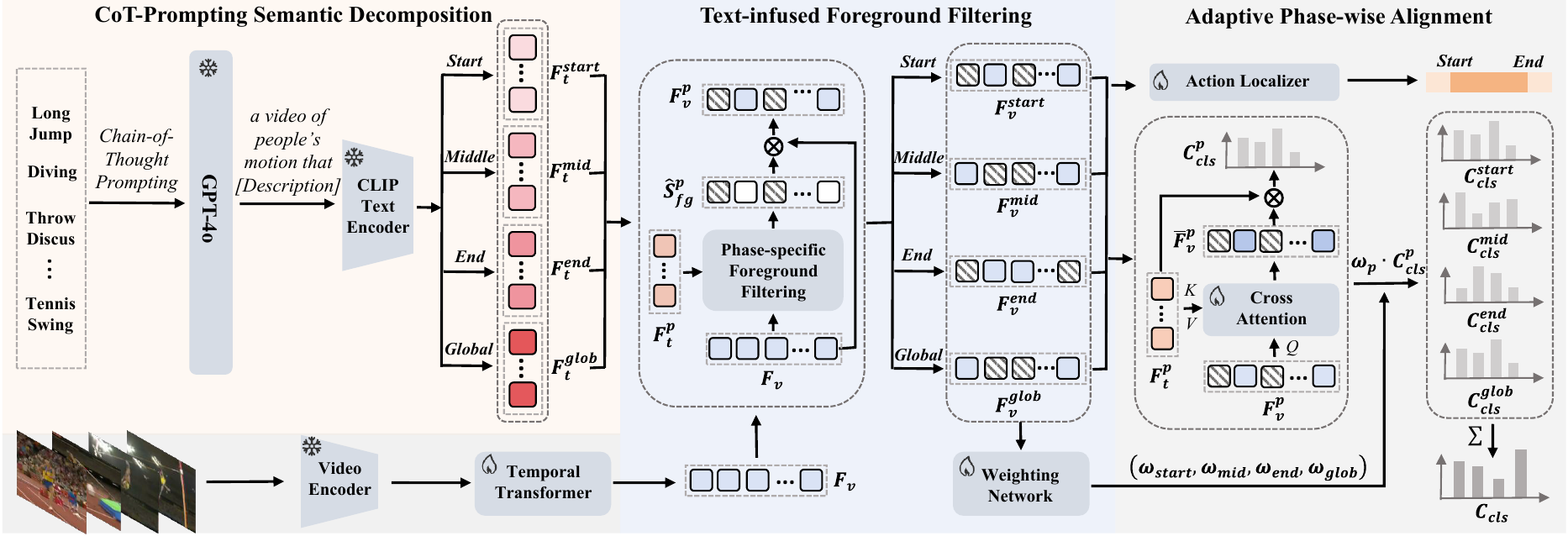}
  \caption{
  An overview of the proposed framework, which comprises three key modules: CoT-Prompting Semantic Decomposition (CSD), Text-infused Foreground Filtering (TIF) and Adaptive Phase-wise Alignment (APA). 
  Specifically, the CSD module temporally decomposes action labels into multiple phase descriptions $F_{t}^{p}$, where {$p \in \{start, mid, end, glob\}$}. 
  The TIF module then leverages phase-specific semantic cues to adaptively filter
  action-relevant segments for each phase, yielding phase-specific visual representations $F_{v}^{p}$. 
  Finally, the APA module performs phase-wise alignment and adaptively aggregates the alignment results for final action detection.
  }
 \label{framework}
 \vspace{-0.5em}
\end{figure*}

\noindent
\textbf{Chain-of-Thought Prompting} has recently emerged as a powerful paradigm that enables LLMs to perform complex reasoning by generating interpretable intermediate steps~\cite{wang2022self, wei2022chain, gao2023pal, maniparambil2023enhancing, yu2025gcot}. 
By explicitly decomposing a problem into a sequence of logically connected reasoning steps, it allows emulation of human-like analytical processes, thereby improving performance in tasks that demand structured reasoning~\cite{feng2023towards, yao2023tree}. 
Motivated by this, we extend CoT prompting to action understanding and explore its potential for temporal action decomposition. 
Rather than trivially applying CoT to textual reasoning~\cite{turpin2023language, diao2024active}, we leverage it to decompose an action label into multiple semantically coherent phase descriptions. 
This progressive reasoning process mirrors human cognition, where understanding an action unfolds step by step, that each phase is logically dependent on previous one. 
Such structured decomposition enables the generation of temporally consistent and semantically rich phase representations, providing reliable semantics for transferable action pattern learning.

%% file: sec/3_methodology.tex
\section{Methodology}

\begin{table*}
   \centering
    \caption{Prompt template for guiding GPT-4o to generate phase-specific and global descriptions of action labels and illustrative examples.  The \textcolor{blue}{“Start”} and \textcolor{blue}{“Middle”} phases of LongJump and PoleVault exhibit semantic similarity.}
  \resizebox{\textwidth}{!}{
  \begin{tabular}{c|c|c|c|c}
   \hline
    \multirow{3}{*}{\textbf{Action}} & \multicolumn{3}{c|}{\textbf{Prompt: Decompose the action of $\langle Action \rangle$ into coherent three phases based on the }} & \multirow{3}{*}{\textbf{\shortstack{Prompt: Describe how \\ a person does $\langle Action \rangle$.}}} \\
    & \multicolumn{3}{c|}{\textbf{natural temporal progression of the action. Please provide the output step by step.}} & \\
    \cline{2-4}
    & \textcolor{blue}{\textbf{Start}} & \textcolor{blue}{\textbf{Middle}} & \textbf{End} & \\
    \hline
    \hline
    \multirow{2}{*}{LongJump} & \textit{The person would \textcolor{blue}{run down}} & \textit{The person would plant one foot} & The person would extend their & The person would sprint down the track \\
    & \textcolor{blue}{\textit{the track to gain speed.}} & \textit{and \textcolor{blue}{push off the ground}.} & legs and land in the sand. & and jump forward into the sandpit. \\
    \hline
    \multirow{2}{*}{PoleVault} & \textit{The person would \textcolor{blue}{run down}} & \textit{The person would plant the pole} & The person would clear the bar & The person would sprint down the track, \\
    & \textcolor{blue}{\textit{the track to gain speed.}} & \textit{and \textcolor{blue}{push off the ground}.} & and fall onto the landing mat. & vault with a pole, and clear a high bar. \\
    \hline
  \end{tabular}
  }
 \label{Prompt template}
\end{table*}

\subsection{Problem Definition}
Given a training set of untrimmed videos $D_{train}$, each video is represented as a sequence of visual features $X=\{x_t\}_{t=1}^{T}$, where $T$ denotes snippets (a few sequences of frames).  
The corresponding annotations are defined as $Y=\{s_n, e_n, c_n\}^{N}_{n=1}$, where $s_n$ and $e_n$ indicate the start and end points of the $n$-th action, and $c_n \in \mathcal{C}_{train}$ denotes the action categories for training.
In the open-vocabulary setting, the label sets for training and testing are disjoint, i.e., $\mathcal{C}_{train} \cap \mathcal{C}_{test} = \emptyset$.
The goal of Open-Vocabulary Temporal Action Detection (OV-TAD) is to localize and classify actions from unseen classes in untrimmed test videos, by leveraging transferable knowledge learned from the seen categories during training.

\subsection{CoT-Prompting Semantic Decomposition}
Previous methods encode action labels directly as semantic representations, which fails to effectively capture fine-grained semantic similarities, limiting the efficacy under open-vocabulary setting. 
Intuitively, while certain actions may exhibit obvious semantic differences at label-level, they still share similar temporal segments, which can serve as shared prior that is transferable across different actions. 
Intuitively, this observation aligns with human cognition, where action understanding involves perceiving multiple temporal phases as distinct yet coherent components.

Recently, the Chain-of-Thought (CoT) reasoning capability of large language models (LLMs)~\cite{wei2022chain} has been explored to enable step-by-step reasoning, that reflects the temporal progression inherent in human action perception. 
Motivated by this, we leverage the CoT reasoning to automatically decompose each action label into coherent temporal phases. 
In this work, we adopt a three-phase decomposition, i.e., start, middle, and end, to extract transferable semantic knowledge shared across semantically diverse actions. 
Additionally, we include a holistic action description, rather than the raw label itself, to provide complementary global context.
Unlike conventional label-level representations, this phase-aware semantic representations guide the learning of transferable action patterns, thus achieving more effective cross-action generalization in OV-TAD.

Table~\ref{Prompt template} shows the prompt template used to guide the CoT reasoning process for coherent phase-wise action decomposition with GPT-4o~\cite{hurst2024gpt} as the LLM backbone. 
Based on the generated phase-wise descriptions, we construct standardized prompts in the form of `a video of people’s motion that [Description]', which are encoded using a pre-trained CLIP text encoder $f_{text}$ to obtain the corresponding embeddings.
Specifically, for label $c\in \mathcal{C}_{train} \cup \mathcal{C}_{test}$, GPT-4o generates a set of phase-wise descriptions $D_{c}=\{d^{p}_{c} | p\in \mathcal{P}\}$, where $\mathcal{P} = \{start, middle, end, glob\}$. 
Each description $d^{p}_{c}$ is encoded as $s^{p}_{c}=f_{text}(d^p_c)$. 
We apply phase-specific encoders $\Phi_{txt}$ to map textual features into shared representation space as:
\begin{equation}
t^{p}_{c} = \Phi_{txt}(s^p_c) \in \mathbb{R}^{D},
\end{equation}
where $D$ denotes the feature dimension. 



\subsection{Text-infused Foreground Filtering}
Following~\cite{lee2024text}, we first extract the initial video features with visual encoder $\Phi_{vis}(\cdot)$, which are then processed through the temporal transformer (a stack of $L$ layers). 
Each layer involves a multi-head self-attention (MHSA) followed by a feed-forward network (FFN).
Thus, we obtain the final representation of ${F}_{v} \in \mathbb{R}^{T \times D}$, where $T$ denotes the number of temporal snippets. 

Foreground filtering~\cite{lee2024text} aims to suppress irrelevant background information and enhance the model’s focus on foreground action segments. 
To improve the semantic alignment with phase-wise descriptions, we extend it to phase-specific foreground extraction, highlighting relevant segments for each phase. A straightforward approach is to partition the video into predefined temporal segments, each corresponding to one phase—termed \textit{static foreground filtering}. 
However, real-world videos typically contain multiple actions with variable durations, rendering such fixed segmentation is insufficient to robustly align visual features with their corresponding phase descriptions.
To overcome this, we propose a Text-infused Foreground Filtering (TIF) module that leverages phase-wise semantic cues to adaptively filter action-relevant segments for each phase. 

Specifically, for each phase $p \in \mathcal{P}$, we compute the similarity between the visual features ${F}_{v}$ and the corresponding phase-$p$ textual embeddings. At each timestep, we select the maximum similarity across all classes as the foreground confidence score:
\begin{equation}
 S^{p}_{fg} = Softmax({\underset{C}{\max}}({F}_{v} F_{t}^{p \top})) \in \mathbb{R}^{T},
\end{equation}
where $F_{t}^{p}=\{t^{p}_{c}\}_{c=1}^{C} \in \mathbb{R}^{C \times D}$ denotes the set of phase-$p$ semantic embeddings for the $C$ training classes, and Softmax ensures the normalized range within $[0, 1]$.

The resulting foreground probability sequence $S^{p}_{fg}$ provides a confidence score for at each time step, indicating the probability of action occurrence in phase $p$. We then binarized $S^{p}_{fg}$ to $\hat{S}^{p}_{fg}$ with a threshold defined as the average similarity over all temporal positions. The binary mask $\hat{S}^{p}_{fg}$ is subsequently applied to selectively filter ${F}_{v}$, yielding the phase-specific visual representation ${F}_{v}^{p}$:
\begin{equation}
 {F}_{v}^{p} = \hat{S}_{fg}^{p} \cdot {F}_{v} \in \mathbb{R}^{T}.
\end{equation}

\subsection{Adaptive Phase-wise Alignment}
Once obtaining the visual and textual features, an intuitive solution is to conduct naive \textit{global alignment} between these two modalities for action detection.
Specifically, a unified textual representation is obtained by \textit{concatenating} the phase-specific descriptions, while the visual representation is derived by \textit{average pooling} over all snippet features. 
Action classification is then performed by computing the semantic similarity between these global representations, followed by the subsequent action localization for start and end boundaries.
However, this global alignment often exhibit instability in capturing the fine-grained phase-level action semantics that transfer from seen to unseen labels. 

To address this, we propose a novel Adaptive Phase-wise Alignment strategy. 
On one hand, \textit{Phase-wise} denotes we conduct the text-video alignment in a phase-wise manner, which is consistent with the obtained phase-level textual descriptions and filtered visual features.
While \textit{Adaptive} indicates we iteratively aggregate the alignment results across different phases to produce the final prediction.

\noindent
\textit{\textbf{Action Classification}.} 
Compared to global alignment, our method enables finer correspondence with textual descriptions by focusing on phase-specific action segments.
Specifically, for each phase $p \in \mathcal{P}$, we first leverage cross-attention mechanism to infuse text information into phase-specific video, resulting in text-aware visual representation:
\begin{equation}
{\bar{F}}^{p}_{v} = Softmax(\frac{Q({F}^{p}_{v})K(F_{t}^{p})^{\top}}{\sqrt{D}})V(F_{t}^{p}).
\end{equation}

The classification score is then computed via similarity between refined visual feature and the textual features of all action categories: 
\begin{equation}
C_{cls}^{p} = {\bar{F}}^{p}_{v} F_{t}^{p \top} \in \mathbb{R}^{T \times C},
\end{equation}
where $C_{cls}^{p}$ represents the probability over all categories at time step $t$ within phase $p$.

A common approach to obtaining the final classification result is to aggregate phase-wise scores via average pooling.  
However, this method fails to adaptively adjust each phase’s contribution based on its relevance. 
Instead, we propose an adaptive aggregation strategy, where phase-specific weights are predicted as:
\begin{equation}
\omega_{p} = Sigmoid(W_{p}({F}_{v}^{p})),
\end{equation}
where $W_{p}$ is a weighting network.
The final prediction is computed as weighted sum of all phases:
\begin{equation}
 C_{cls} = \underset{p \in \mathcal{P}}\sum (\omega_{p} \cdot C_{cls}^{p}) \in \mathbb{R}^{T \times C}.
\end{equation}

Compared to intuitive average pooling, our adaptive aggregation offers more flexibility by enabling the model to emphasize more discriminative phases while down-weighting less informative ones.
Finally, we employ the cross-entropy loss for action classification as:
\begin{equation}
 \mathcal{L}_{cls} = CrossEntropy(C_{cls}, G_{cls}),
\end{equation}
where $G_{cls} \in \mathbb{R}^{T \times C}$ is the one-hot ground-truth label.

\noindent
\textit{\textbf{Action Localization}} focuses on the unified representation by concatenating visual features from all phases along the feature dimension, followed by a linear projection $L$ to match the original feature space:
\begin{equation}
\hat{F}_{v} = L([{F}_{v}^{start};{F}_{v}^{mid};{F}_{v}^{end};{F}_{v}^{glob}]) \in \mathbb{R}^{T},
\end{equation}
where $[\cdot;\cdot;\cdot]$ denotes concatenation along the feature dimension, $L$ is implemented by multi-layer MLP.
$\hat{F}_{v}$ is then fed into a foreground-aware head and a regression head respectively, predicting the distances $d_t^s$ and $d_t^e$ from each time step $t$ to the action start and end boundaries. 
These predictions are supervised using a foreground-aware loss $\mathcal{L}_{fg}$ and a DIoU-based localization loss $\mathcal{L}_{loc}$ as in~\cite{lee2024text}.

\subsection{Training and Inference}
To sum up, combining the action classification loss $\mathcal{L}_{cls}$, the foreground-aware loss $\mathcal{L}_{fg}$ and the action localization loss $\mathcal{L}_{loc}$ together, the overall objective is formulated as:
\begin{equation}
\mathcal{L} =\mathcal{L}_{cls} + \mathcal{L}_{fg} + \mathcal{L}_{loc},
\end{equation}

At test time, we first utilize LLM to decompose each action label from the test split into multi-phase textual descriptions.  
Subsequently, for each time step $t$ in test video, the model predicts $(d_t^s, d_t^e, p(c_t))$ via phase-wise classification and localization. 
Here, $d_t^s$ and $d_t^e$ denote estimated distances from $t$ to start and end boundaries of action instance, respectively, while $p(c_t)$ represents confidence score of predicted action. 
Finally, redundant proposals are suppressed with SoftNMS~\cite{bodla2017soft} yielding final action predictions.

%% file: sec/4_experiment.tex
\section{Experiments}

\begin{table*}[t]
\centering
\caption{ Performance comparison with the state-of-the-art methods on THUMOS14 and ActivityNet v1.3. * indicates closed-set TAD methods adapted to the OV-TAD setting. The best and second-best results are highlighted in \textbf{\textcolor{red}{Red}} and \textbf{\textcolor{blue}{Blue}}, respectively.}
\resizebox{\textwidth}{!}{
\begin{tabular}{c c c c c c c c c c c c} 
\hline
\multirow{2}{*}{Data Split} & \multirow{2}{*}{Methods} & \multicolumn{6}{c}{THUMOS14} & \multicolumn{4}{c}{ActivityNet v1.3} \\
\cmidrule(lr){3-8} \cmidrule(lr){9-12}
 &  & 0.3 & 0.4 & 0.5 & 0.6 & 0.7 & Avg mAP(\%) & 0.5 & 0.75 & 0.95 & Avg mAP(\%) \\
\hline
\multirow{11}{*}{\shortstack{50\% Seen \\ 50\% Unseen}} 
& TriDet* (ICCV'23) & 15.2 & 13.2 & 10.8 & 7.9 & 5.2 & 10.5 & 19.1 & 11.5 & 1.1 & 11.4 \\
& DyFADet* (ECCV'24) & 17.5 & 14.9 & 12.2 & 9.2 & 5.7 & 11.9 & 23.8 & 14.2 & 1.8 & 13.6 \\
& DiGIT* (CVPR'25) & 19.1 & 16.2 & 13.5 & 10.3 & 6.1 & 13.0 & 27.5 & 17.3 & 2.3 & 16.0 \\
& EffPrompt (ECCV'22) & 37.2 & 29.6 & 21.6 & 14.0 & 7.2 & 21.9 & 32.0 & 19.3 & 2.9 & 19.6 \\
& STALE (ECCV'22) & 38.3 & 30.7 & 21.2 & 13.8 & 7.0 & 22.2 & 32.1 & 20.7 & 5.9 & 20.5 \\
& DeTAL (TPAMI'24) & 38.3 & 32.3 & 24.4 & 16.3 & 9.0 & 24.1 & 34.4 & 23.0 & 4.0 & 22.4 \\
& CSP (JCST'25) & 41.2 & 33.4 & 24.8 & 17.3 & 10.9 & 25.5 & 38.4 & 26.4 & 5.2 & 25.7 \\
& ZEETAD (WACV'24) & 45.2 & 38.8 & 30.8 & 22.5 & 13.7 & 30.2 & 39.2 & 25.7 & 3.1 & 24.9 \\
& STOV (WACV'25) & 56.3 & - & 34.4 & - & 11.3 & 34.0 & 48.4 & 28.7 & - & 27.9 \\
& Ti-FAD (NeurlPS'24) & \textcolor{blue}{\underline{57.0}} & \textcolor{blue}{\underline{51.4}} & \textcolor{blue}{\underline{43.3}} & \textcolor{blue}{\underline{33.0}} & \textcolor{blue}{\underline{21.2}} & \textcolor{blue}{\underline{41.2}} & \textcolor{blue}{\underline{50.6}} & \textcolor{blue}{\underline{32.2}} & \textcolor{blue}{\underline{5.2}} & \textcolor{blue}{\underline{32.0}} \\
\cmidrule(lr){2-12}
& \textbf{Ours} & \textcolor{red}{\textbf{65.4}} & \textcolor{red}{\textbf{57.2}} & \textcolor{red}{\textbf{49.7}} & \textcolor{red}{\textbf{37.9}} & \textcolor{red}{\textbf{24.3}} & \textcolor{red}{\textbf{46.9}} & \textcolor{red}{\textbf{53.1}} & \textcolor{red}{\textbf{35.3}} & \textcolor{red}{\textbf{7.7}} & \textcolor{red}{\textbf{34.6}}\\
\hline
\multirow{11}{*}{\shortstack{75\% Seen \\ 25\% Unseen}} 
& TriDet* (ICCV'23) & 25.9 & 22.5 & 18.2 & 13.1 & 6.2 & 17.2 & 25.5 & 15.2 & 2.0 & 15.3 \\
& DyFADet* (ECCV'24) & 27.6 & 23.9 & 19.4 & 13.8 & 6.7 & 18.3 & 28.9 & 17.6 & 2.5 & 17.1 \\
& DiGIT* (CVPR'25) & 29.0 & 25.2 & 20.5 & 14.3 & 7.0 & 19.2 & 32.2 & 19.7 & 2.9 & 18.8 \\
& EffPrompt (ECCV'22) & 39.7 & 31.6 & 23.0 & 14.9 & 7.5 & 23.3 & 37.6 & 22.9 & 3.8 & 23.1 \\
& STALE (ECCV'22) & 40.5 & 32.3 & 23.5 & 15.3 & 7.6 & 23.8 & 38.2 & 25.2 & 6.0 & 24.9 \\
& DeTAL (TPAMI'24) & 39.8 & 33.6 & 25.9 & 17.4 & 9.9 & 25.3 & 39.3 & 26.4 & 5.0 & 25.8 \\
& CSP (JCST'25) & 42.7 & 35.5 & 26.4 & 18.5 & 12.0 & 27.0 & 41.1 & 28.8 & 7.4 & 28.1 \\
& ZEETAD (WACV'24) & 61.4 & 53.9 & 44.7 & 34.5 & 20.5 & 43.2 & 51.0 & 33.4 & 5.9 & 32.5 \\
& STOV (WACV'25) & 59.5 & - & 37.5 & - & 12.5 & 36.9 & 52.0 & 30.6 & - & 30.1 \\
& Ti-FAD (NeurlPS'24) & \textcolor{blue}{\underline{64.0}} & \textcolor{blue}{\underline{58.5}} & \textcolor{blue}{\underline{49.7}} & \textcolor{blue}{\underline{37.7}} & \textcolor{blue}{\underline{24.1}} & \textcolor{blue}{\underline{46.8}} & \textcolor{blue}{\underline{53.8}} & \textcolor{blue}{\underline{34.8}} & \textcolor{blue}{\underline{7.0}} & \textcolor{blue}{\underline{34.7}} \\
\cmidrule(lr){2-12}
& \textbf{Ours} & \textcolor{red}{\textbf{70.5}} & \textcolor{red}{\textbf{63.8}} & \textcolor{red}{\textbf{54.6}} & \textcolor{red}{\textbf{43.1}} & \textcolor{red}{\textbf{28.3}} & \textcolor{red}{\textbf{52.1}} & \textcolor{red}{\textbf{56.2}} & \textcolor{red}{\textbf{37.8}} & \textcolor{red}{\textbf{8.6}} & \textcolor{red}{\textbf{37.4}}\\
\hline
\end{tabular}
}
\label{performance comparison}
\end{table*}

\begin{table*}[t!]
  \normalsize
  \centering
  \caption{Analysis of different action phase number on THUMOS14 under the 50\% seen / 50\% unseen split. The phase number adopted in this paper is highlighted with a \colorbox{blue!10}{blue background}.}
  \begin{tabularx}{0.9\textwidth}{ >{\raggedright\arraybackslash}p{0.35\linewidth}  >{\centering\arraybackslash}X  >{\centering\arraybackslash}X  >{\centering\arraybackslash}X  >{\centering\arraybackslash}X
  >{\centering\arraybackslash}p{0.1\linewidth}} 
  \hline
   \multirow{2}{*}{\textbf{Phase Number}} & \multicolumn{4}{c}{mAP@tIOU (\%)} & \multirow{2}{*}{Time (s)} \\
   \cmidrule(lr){2-5} 
   & 0.3 & 0.5 & 0.7 & Avg & \\
   \hline
    One \textit{(Glob)} & 59.3 & 45.5 & 21.9 & 42.5 & 27.6 \\
    Two \textit{(Start, End)} & 61.0 & 46.9 & 22.8 & 44.0 & 29.2 \\
    Three \textit{(Start, Mid, End)} & 63.9 & 48.4 & 23.6 & 45.3 & 30.5 \\
    \rowcolor{blue!10} Four \textit{(Start, Mid, End, Glob)} & 65.4 & 49.7 & 24.3 & 46.9 & 32.4 \\
    Five \textit{(Start, Mid1, Mid2, End, Glob)} & 66.3 & 50.4 & 24.8 & 47.6 & 34.8 \\
    Six \textit{(Start, Mid1, Mid2, Mid3, End, Glob)} & 66.7 & 50.6 & 25.1 & 47.8 & 37.3\\
    \hline
  \end{tabularx}
  \label{Phase Number THUMOS}
\end{table*}

\subsection{Experimental Details}
\textbf{Datasets.} 
We evaluate our method on two standard Temporal Action Detection (TAD) benchmarks: ActivityNet v1.3~\cite{caba2015activitynet}, with 19,994 untrimmed videos from 200 classes, and THUMOS14~\cite{idrees2017thumos}, containing 200 validation and 213 test videos across 20 categories.
Following~\cite{ju2022prompting}, we adopt two open-vocabulary splits: training on 75\% / 50\% of the classes and testing on the remaining 25\% / 50\%, each averaged over 10 random splits for robustness.

\noindent
\textbf{Evaluation Metric.}
We adopt mean Average Precision (mAP) as the evaluation metric computed by averaging precision across multiple temporal Intersection over Union (tIoU) thresholds. For THUMOS14, tIoU thresholds range from 0.3 to 0.7 with a step size of 0.1, while for ActivityNet v1.3, they span from 0.5 to 0.95 with a step size of 0.05.

\noindent
\textbf{Implementation Details.} 
For fair comparison with prior TAD methods, we follow~\cite{nag2022zero,lee2024text} and adopt two-stream I3D features as input. Our model is trained for 12 epochs on THUMOS14 and 7 epochs on ActivityNet v1.3 using the Adam optimizer. All experiments are conducted on a single NVIDIA A100 GPU. 

\noindent
\textbf{Baselines.} 
We compare our method with ten state-of-the-art TAD approaches, including three closed-set methods (TriDet~\cite{shi2023tridet}, DyFADet~\cite{yang2024dyfadet}, DiGIT~\cite{kim2025digit}) and seven Open-Vocabulary methods (EffPrompt~\cite{ju2022prompting}, STALE~\cite{nag2022zero}, DeTAL~\cite{li2024detal}, CSP~\cite{wangconcept}, ZEETAD~\cite{phan2024zeetad}, STOV~\cite{hyun2025exploring}, Ti-FAD~\cite{lee2024text}). 
Closed-set methods are adapted to the OV-TAD setting following the protocol in~\cite{li2024detal,wangconcept}, which splits the action label space into disjoint training and testing subsets. 
These baselines form a comprehensive benchmark to assess the effectiveness of our approach.

\subsection{Comparison with State-of-the-Arts}
We evaluate our method against state-of-the-art OV-TAD approaches as well as adapted fully-supervised TAD baselines. 
As shown in Table~\ref{performance comparison}, our method consistently outperforms existing methods on both THUMOS14 and ActivityNet1.3 datasets, achieving highest mAP scores across all tIoUs. 
For example, under the 50\% Seen / 50\% Unseen split, our method achieves a 13.8\% and 8.1\% relative lift in average mAP over SOTA competitor Ti-FAD on two benchmarks, respectively. 
This is consistent with the 75\% Seen / 25\% Unseen split, clearly demonstrating the effectiveness of our design. 
Specifically, the CSD module leverages LLMs’ CoT reasoning to decompose action labels into multi-phase descriptions, while the TIF and APA modules enable adaptive phase-wise visual–textual alignment. 

\subsection{Ablation Studies}

\noindent
\textbf{Analysis of different phase number.} 
We conduct experiments with different decomposition phases, ranging from a global description to six phases per action. 
As shown in Table~\ref{Phase Number THUMOS}, performance consistently improves with more phases, indicating that richer phase semantics enhance the model’s transferability to unseen actions. 
However, the improvement becomes marginal when the number exceeds four. 
This may attribute to the saturation of informative semantics and the increasing of noisy descriptions. Additionally, increasing the number of phases leads to more time cost owing to the additional alignment computations.
For trade-off between accuracy and efficiency, we adopt (start, middle, end, global) in our implementation. 

\noindent
\textbf{Analysis of different LLM backbones.}
To evaluate the robustness across LLM backbones, we compare four widely adopted LLMs: Qwen3~\cite{yang2025qwen3}, DeepSeek-v3~\cite{liu2024deepseek}, GPT-4~\cite{achiam2023gpt}, and GPT-4o~\cite{hurst2024gpt}. 
Table~\ref{LLM Backbone main} shows that the overall performance remains largely consistent across different backbones. 
The marginal variations among these backbones indicate that our model is robust to the choice of LLM, and does not depend on specific backbone, which is desirable for deployment. 

\begin{table}[t!]
  \normalsize
   \caption{Analysis of different LLM backbones on THUMOS14 under the 50\% seen / 50\% unseen split.}
  \begin{tabularx}{\columnwidth}{ >{\raggedright\arraybackslash}p{0.3\linewidth}  >{\centering\arraybackslash}X  >{\centering\arraybackslash}X  >{\centering\arraybackslash}X  >{\centering\arraybackslash}X}
  \hline
   \multirow{2}{*}{LLM Backbone} & \multicolumn{4}{c}{mAP@tIOU (\%)} \\
   \cmidrule(lr){2-5}
   & 0.3 & 0.5 & 0.7 & Avg \\
   \hline
    Qwen3 & 64.9 & 49.1 & 24.0 & 46.2\\
    Deepseek v3 & 64.5 & 49.0 & 23.7 & 46.2\\
    GPT-4 & 65.1 & 49.4 & 24.1 & 46.6 \\
    GPT-4o & \textcolor{red}{\textbf{65.4}} & \textcolor{red}{\textbf{49.7}} & \textcolor{red}{\textbf{24.3}} & \textcolor{red}{\textbf{46.9}} \\
    \hline
  \end{tabularx}
  \label{LLM Backbone main}
\end{table}

\begin{table}[t!]
  \normalsize
  \caption{Analysis of each components on THUMOS14.}
  \begin{tabularx}{\columnwidth}{ >{\raggedright\arraybackslash}p{0.13\linewidth}  >{\centering\arraybackslash}X  >
  {\centering\arraybackslash}X  >
  {\centering\arraybackslash}X  >{\centering\arraybackslash}p{0.2\linewidth}  >{\centering\arraybackslash}p{0.2\linewidth}}
  \hline
   \multirow{2}{*}{Method} & \multirow{2}{*}{CSD} & \multirow{2}{*}{TIF} & \multirow{2}{*}{APA} & \multicolumn{2}{c}{mAP@AVG} \\
    \cmidrule(lr){5-6}
   & & & & 50\%-50\% & 75\%-25\% \\
   \hline
   Baseline & \ding{55} & \ding{55} & \ding{55} & 40.3 & 45.9 \\
   \hline
   \multirow{3}{*}{Ours} & \ding{51} & \ding{55} & \ding{55} & 42.1 & 47.8\\
   & \ding{51} & \ding{51} & \ding{55} & 43.6 & 49.0\\
   & \ding{51} & \ding{51} & \ding{51} & \textcolor{red}{\textbf{46.9}} & \textcolor{red}{\textbf{52.1}}\\
  \hline
  \end{tabularx}
  \label{component analysis}
\end{table}

\noindent
\textbf{Analysis of each component.}  
Table~\ref{component analysis} shows the ablations on three core modules: CSD, TIF and APA. 
In the baseline model (Row 1), label-level textual features are directly used as semantic representations and aligned with averaged visual features. 
Rows 2, 3 and 4 progressively incorporate CSD, TIF and APA to assess their individual contributions.
Clearly, Row 2 decomposes action labels into multi-phase descriptions via CoT reasoning, improves the final performance under both splits. 
This is reasonable since compared to encoding coarse label-level semantics, the CSD module provides more fine-grained, phase-specific descriptions, capturing richer semantic nuances.
The performance gain of Row 3 and 4 also justifies that aligning phase-specific visual features with corresponding textual semantics enables learning of transferable, phase-level action patterns.


\noindent
\textbf{Analysis of Text-infused Foreground Filtering (TIF).}
We compare with three devised variants of TIF module: 
\textbf{1)} \textit{w/o Filtering} removes the foreground filtering step. 
\textbf{2)} \textit{Static Filtering} divides the video into fixed segments along the time sequence, each treated as a distinct phase for matching. 
\textbf{3)} \textit{Text-infused Foreground Filtering} leverages phase-specific semantics to adaptively filter action-relevant segments for each phase. 
As in Table~\ref{visual refinement}: 
1) Filtering method improves performance over the non-filtering baseline, indicating that emphasizing action-relevant snippets benefits action detection. 
2) Text-infused filtering outperforms the static one, demonstrating its effectiveness in capturing phase-aligned visual cues, leading to more accurate cross-modal alignment.

\begin{table}[t!]
  \normalsize
    \caption{Analysis of Text-infused Foreground Filtering on THUMOS14 under the 50\% seen / 50\% unseen split.}
  \begin{tabularx}{\columnwidth}{ >{\raggedright\arraybackslash}p{0.4\linewidth}  >{\centering\arraybackslash}X  >{\centering\arraybackslash}X  >{\centering\arraybackslash}X  >{\centering\arraybackslash}X}
  \hline
   \multirow{2}{*}{Method} & \multicolumn{4}{c}{mAP@tIOU (\%)} \\
   \cmidrule(lr){2-5}
   & 0.3 & 0.5 & 0.7 & Avg \\
   \hline
    w/o Filtering & 62.1 & 46.7 & 21.9 & 44.3\\
    Static Filtering & 63.4 & 48.0 & 22.8 & 45.5\\
    \textbf{Text-infused Filtering} & \textcolor{red}{\textbf{65.4}} & \textcolor{red}{\textbf{49.7}} & \textcolor{red}{\textbf{24.3}} & \textcolor{red}{\textbf{46.9}} \\
    \hline
  \end{tabularx}
  \label{visual refinement}
\end{table}

\begin{table}[t!]
  \normalsize
   \caption{Analysis of Adaptive Phase-wise Alignment on THUMOS14 under the 50\% seen / 50\% unseen split. “Global” represents global alignment between text features and averaged visual features,  “Phase-wise” represents phase-wise alignment followed by aggregation.}
  \begin{tabularx}{\columnwidth}{ >{\raggedright\arraybackslash}p{0.2\linewidth}  >{\centering\arraybackslash}p{0.16\linewidth}  >{\centering\arraybackslash}X  >{\centering\arraybackslash}X  >{\centering\arraybackslash}X  >{\centering\arraybackslash}X}
  \hline
   \multirow{2}{*}{Alignment} & \multirow{2}{*}{Method} & \multicolumn{4}{c}{mAP@tIOU (\%)} \\
   \cmidrule(lr){3-6}
   & & 0.3 & 0.5 & 0.7 & Avg \\
  \hline
  \multirow{6}{*}{Global} & Label & 56.2 & 42.7 &  20.4 & 40.3 \\
  \cmidrule(lr){2-6}
   &  Start & 57.4 & 43.8 & 21.1 & 41.2\\
   &  Middle & 58.1 & 44.2 & 21.4 & 41.5 \\
   &  End & 57.8 & 44.6 & 21.5 & 41.7\\
   &  Global & 59.3 & 45.5 & 21.9 & 42.5\\
   & Merge & 61.1 & 47.3 & 22.7 & 44.0\\
   \hline
   \multirow{2}{*}{Phase-wise} & Average & 63.6  & 48.5 & 23.1 & 45.8\\
   & \textbf{Adaptive} & \textcolor{red}{\textbf{65.4}} & \textcolor{red}{\textbf{49.7}} & \textcolor{red}{\textbf{24.3}} & \textcolor{red}{\textbf{46.9}} \\
    \hline
  \end{tabularx}
  \label{phase-wise alignment}
\end{table}

\noindent
\textbf{Analysis of Adaptive Phase-wise Alignment (APA).}
We further compare with two alignment paradigms: 
\textbf{1)} \textit{Global} matches averaged visual features with (a) label-level semantic embeddings, (b) phase-specific semantic descriptions, and a Merge variant that fuses all phase descriptions into a single representation, individually. 
\textbf{2)} \textit{Phase-wise} conducts phase-wise alignment and aggregates alignments via (a) simple averaging, and (b) adaptive weighting based on dynamic phase importance. 
Table~\ref{phase-wise alignment} shows that: 
1) Phase-wise strategies consistently outperform global ones, confirming that the performance gains arise from the \textbf{combination} of action decomposition and adaptive phase-wise alignment, rather than richer textual semantics alone. 
2) Adaptive aggregation surpasses simple averaging by emphasizing discriminative phases and suppressing less informative ones.

\begin{figure*}[t]
\centering
  \includegraphics[width=\textwidth]{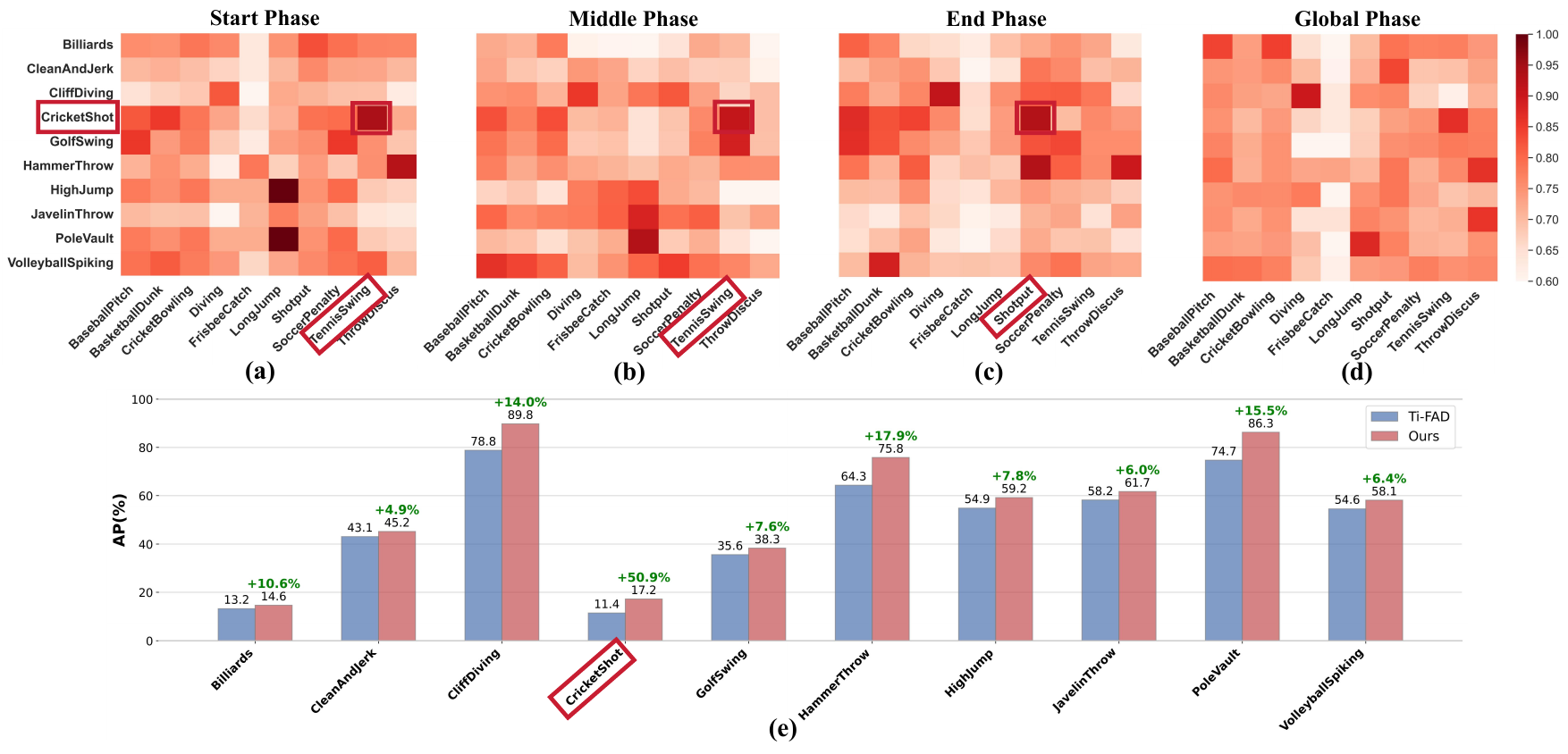}
  \caption{Phase-wise Semantic Similarity ((a)-(d)) and Per-unseen class AP (\%) (e) at tIoU threshold 0.5 on THUMOS14 under the 50\% seen / 50\% unseen split. For (a)-(d), the vertical denotes unseen (testing) classes, the horizontal denotes seen (training) classes. }
  \label{AP1}
\end{figure*}

\begin{figure*}[t]
\centering
  \includegraphics[width=\textwidth]{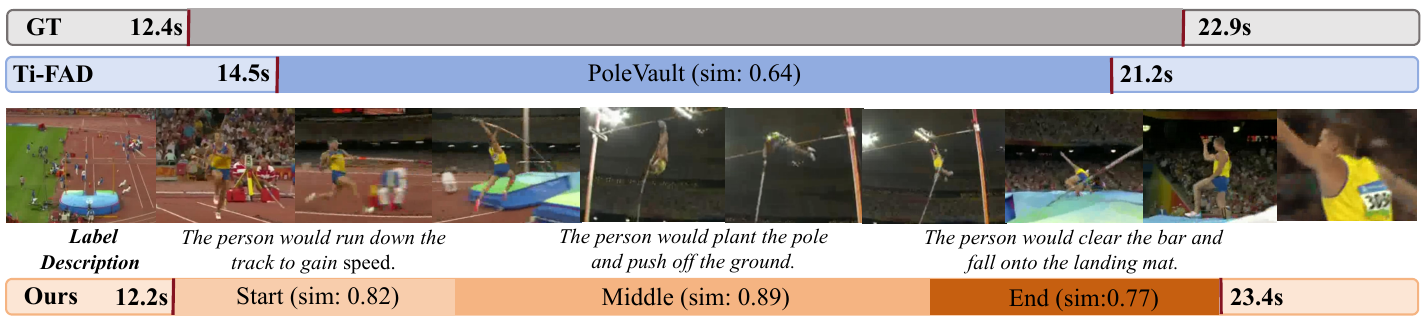}
  \caption{Visualization of the detection results on THUMOS14 under the 50\% seen / 50\% unseen split. “sim” represents the visual-textual similarity at each phase.}
  \label{localization}
\end{figure*}

\noindent
\textbf{Analysis of Phase-wise Semantic Similarity.} 
We visualize the similarity between training and testing label descriptions. 
As in Figure~\ref{AP1} (a)-(d), even semantically distinct actions might share highly consistent phase-specific descriptions (darker red).
For instance, unseen CricketShot and seen TennisSwing share similar start-phase descriptions: \textit{`The person would ... to ... the ball'}.
As in Figure~\ref{AP1} (e), ours consistently outperforms Ti-FAD, and the improvement is more pronounced for categories that phase-level overlap with training set.
Specifically, unseen CricketShot shows high similarities with both TennisSwing (start and middle phases) and Shotput (end phase), leading to greatest performance gain compared with other actions.

\subsection{Qualitative Results}
Figure~\ref{localization} shows the action localization of unseen class PoleVault. 
Compared to Ti-FAD, our method not only produces higher matching scores for each action phase, but also achieves more precise temporal boundaries.
Besides, the textual descriptions are also in line with corresponding snippets, justifying the efficacy of phase-wise label decomposition and adaptive visual-text alignment.

\section{Conclusion}
In this paper, we propose the Phase-wise Decomposition and Alignment (PDA) framework, which facilitates fine-grained action pattern learning for effective knowledge transfer in OV-TAD task.
We propose the CoT-Prompting Semantic Decomposition (CSD) module, leveraging the chain-of-thought ability of LLMs to progressively unfold actions into coherent phase-level descriptions for transferable semantics.
The Text-infused Foreground Filtering (TIF) module adaptively filters action-relevant segments of each phase for visual-textual alignment, and the Adaptive Phase-wise Alignment (APA) module aims to perform phase-wise matching and dynamic phase integrations. 
Extensive experiments demonstrate that our method achieves state-of-the-art performance on two OV-TAD benchmarks.


\section{Acknowledgment}
This work was supported by the National Key R\&D Program of China under Grant 2022YFB3103500, the Natural Science Foundation of China under Grant No. 62506204, the Zhejiang Provincial Natural Science Foundation of China under Grant No. LQN26F020052, the China Postdoctoral Science Foundation under Grant No. 2025M771689 and the Postdoctoral Innovative Talent Support Program under Grant No. GZC20251164.

%% file: sec/X_suppl.tex
\clearpage
\setcounter{page}{1}
\maketitlesupplementary
In the supplementary material, we provide additional experiments to further substantiate the effectiveness of our proposed method. 
We include more implementation details, covering the extraction processes of visual and textual features, model architecture, and training configurations. 
For our core component, CoT-Prompting Semantic Decomposition (CSD), we conduct comprehensive ablations examining the influence of different prompt templates, varying phase numbers, and alternative commercial LLMs for label decomposition, as well as subjective evaluations of the generated descriptions using both GPT-based and human raters. 
Furthermore, we offer more empirical results demonstrating the plug-and-play flexibility of PDA, analyze the impact of diverse visual and textual backbones, compare against prior LLM-based label expansion strategies, report semantic similarity evaluations on the ActivityNet v1.3 dataset, and provide additional qualitative visualizations.

\section{More Implementation Details}
\label{sec:more implementation details}
Following prior works~\cite{nag2022zero, lee2024text, wangconcept}, we employ a two-stream I3D model and the CLIP~\cite{radford2021learning} model for feature extraction. 

\noindent
\textbf{For visual features}, we concatenate the RGB and optical flow features extracted from the two-stream I3D. 
On THUMOS14, video features are extracted from 16-frame segments using a sliding window with a stride of 4. On ActivityNet v1.3, features are extracted with a stride of 16 and subsequently downsampled to 128 dimensions. 

\noindent
\textbf{For textual features}, we use the frozen pre-trained CLIP text encoder (ViT-B/16 and ViT-L/14 variants).

\noindent
\textbf{For the model architecture}, 
the temporal transformer consists of 6 layers, each comprising a multi-head self-attention (MHSA) followed by a feed-forward network (FFN), with a hidden dimension of 512.
The weighting network $W_p$ is implemented as a lightweight Transformer. We first replicate the input 512-dimensional visual feature into four virtual tokens, each representing a phase, and add four learnable phase embeddings to introduce phase-specific priors. The resulting tokens are processed by a 4-head MHSA layer. A linear projection with a hidden dimension of 1024 maps each token to a scalar, and a softmax operation produces the final 4-dimensional phase-wise weight vector.
The linear projection layer $L$ is implemented as a three-layer MLP with a hidden dimension of 1024.

\noindent
\textbf{During training}, we adopt the Adam optimizer with a linear warm-up for the first 5 epochs. 
The initial learning rate is set to 0.0001. A MultiStepLR scheduler is applied for THUMOS14, while cosine annealing~\cite{loshchilov2016sgdr} is used for ActivityNet v1.3. 
The batch size is set to 16 for ActivityNet v1.3 and 2 for THUMOS14.
The code has been submitted along with this pdf.

\section{More Analysis of CoT-Prompting Semantic Decomposition}
\subsection{Analysis of the Prompt Template Design}
In the main submission, we adopt a single prompt template to guide the chain-of-thought reasoning of GPT-4o for generating both phase-specific and global action descriptions. Here, we examine the robustness of our approach to prompt variations by introducing two additional templates, as summarized in Table~\ref{Tempolate}. Although these prompts differ in surface phrasing, they share the same objective—eliciting coherent phase-aware descriptions or holistic motion summaries for each action.
We evaluate the performance of our method using descriptions generated from each prompt variant. As shown in Table~\ref{Tempolate Result}, all prompt versions yield comparable results, with only minor variations across evaluation metrics.
This observation demonstrates two key aspects of robustness: First, GPT-4o shows strong insensitivity to prompt wording; despite differences in linguistic form, it consistently produces coherent and phase-aligned descriptions that capture the essential characteristics of each action. Second, our model is also robust to variations in the input descriptions themselves. Its performance is largely governed by the CoT-Prompting Semantic Decomposition (CSD) and Adaptive Phase-wise Alignment (APA) modules, which focus on learning transferable temporal patterns through phase-wise alignment rather than relying on subtle textual differences among prompt-generated descriptions.

\begin{table*}[t!]
  \normalsize
   \caption{The variants of prompt templates used to guide GPT-4o in generating phase-specific and global action descriptions. The placeholder $\{text\}$ indicates the content to be filled in by GPT-4o.}
  \begin{tabularx}{\textwidth}{ >{\raggedright\arraybackslash}p{0.1\linewidth}  >{\centering\arraybackslash}X  >
  {\centering\arraybackslash}p{0.75\linewidth} }
  \hline
  \textbf{Description} & \textbf{Type} & \textbf{Prompt} \\
  \hline
  \hline
  \multirow{12}{*}{\textbf{Phase-specific}} & \multirow{2}{*}{\textbf{(a)}} & \textbf{\textit{Question}: Given an action of $\langle Action \rangle$, considering how the activity typically begins, evolves, and concludes. After your reasoning, provide a concise phase-wise summary.} \\
  & &  \textit{Answer}: In the start phase, the person would $\{text\}$. In the middle phase, the person would $\{text\}$. In the end phase, the person would $\{text\}$.  \\
   \cmidrule(lr){2-3}
  & \multirow{2}{*}{\textbf{(b)}} & \textbf{\textit{Question}: For the given action of $\langle Action \rangle$, enumerate it into  chronological sub-events and condense these events into three coherent phase descriptions.} \\
  & &  \textit{Answer}: In the start phase, the person would $\{text\}$. In the middle phase, the person would $\{text\}$. In the end phase, the person would $\{text\}$.   \\
  \cmidrule(lr){2-3}
  & \multirow{2}{*}{\textbf{(c)}} & \textbf{\textit{Question}: Decompose the action of $\langle Action \rangle$ into coherent three phases based on the natural temporal progression of the action. Please provide the output step by step.}\\
  & &  \textit{Answer}: In the start phase, the person would $\{text\}$. In the middle phase, the person would $\{text\}$. In the end phase, the person would $\{text\}$. \\
  \hline
  
  \multirow{6}{*}{\textbf{Global}} & \multirow{2}{*}{\textbf{(a)}} & \textbf{\textit{Question}: Describe the motion of a person does $\langle Action \rangle$.}\\
  & &  \textit{Answer}: The person would $\{text\}$.  \\
   \cmidrule(lr){2-3}
  & \multirow{2}{*}{\textbf{(b)}} & \textbf{\textit{Question}: Describe the motion of a person carries out $\langle Action \rangle$. } \\
  & &  \textit{Answer}: The person would $\{text\}$.  \\
  \cmidrule(lr){2-3}
  & \multirow{2}{*}{\textbf{(c)}} & \textbf{\textit{Question}: Describe how a person does $\langle Action \rangle$.} \\
  & &  \textit{Answer}: The person would $\{text\}$.  \\
  \hline
  \end{tabularx}
  \label{Tempolate}
\end{table*}

\begin{table*}[!]
\centering
\caption{Analysis of the Prompt Template Design. The prompt template adopted in this paper is highlighted with a \colorbox{blue!10}{blue background}.}
  \normalsize
  \begin{tabularx}{\textwidth}{ >{\centering\arraybackslash}p{0.10\linewidth}  >
  {\centering\arraybackslash}p{0.12\linewidth}  >
  {\centering\arraybackslash}X  >
  {\centering\arraybackslash}X  >
  {\centering\arraybackslash}X  >
  {\centering\arraybackslash}X  >
  {\centering\arraybackslash}X  >
  {\centering\arraybackslash}X  >
  {\centering\arraybackslash}X  >
  {\centering\arraybackslash}X 
  }
  \hline
   \multirow{2}{*}{\textbf{Data Split}} & \multirow{2}{*}{\textbf{Prompt Type}} 
  & \multicolumn{4}{c}{\textbf{THUMOS14}} & \multicolumn{4}{c}{\textbf{ActivityNet v1.3}} \\
  \cmidrule(lr){3-6} \cmidrule(lr){7-10}
  & & 0.3 & 0.5 & 0.7 & Avg & 0.5 & 0.75 & 0.95 & Avg \\
  \hline
 \multirow{3}{*}{\shortstack{50\% Seen \\ 50\% Unseen}} & \textbf{(a)} & 64.9 & 49.4 & 24.0 & 46.4 & 52.9 & 35.1 & 7.7 & 34.3 \\
 & \textbf{(b)} & 64.5 & 49.0 & 23.7 & 46.2 & 52.6 & 34.8 & 7.3 & 34.0 \\
 &  \cellcolor{blue!10} \textbf{(c)} & \cellcolor{blue!10} 65.4 & \cellcolor{blue!10} 49.7 & \cellcolor{blue!10} 24.3 & \cellcolor{blue!10} 46.9 & \cellcolor{blue!10} 53.1 & \cellcolor{blue!10} 35.3 & \cellcolor{blue!10} 7.7 & \cellcolor{blue!10} 34.6 \\
 \hline
 \multirow{3}{*}{\shortstack{75\% Seen \\ 25\% Unseen}} & \textbf{(a)} & 69.6 & 53.9 & 27.7 & 51.5 & 55.4 & 37.0 & 8.5 & 36.8 \\
 & \textbf{(b)} & 69.3 & 53.5 & 27.4 & 51.3 & 55.1 & 36.9 & 8.3 & 36.6 \\
 & \cellcolor{blue!10} \textbf{(c)} & \cellcolor{blue!10} 70.5 & \cellcolor{blue!10} 54.6 & \cellcolor{blue!10} 28.3 & \cellcolor{blue!10} 52.1 & \cellcolor{blue!10} 56.2 & \cellcolor{blue!10} 37.8 & \cellcolor{blue!10} 8.6 & \cellcolor{blue!10} 37.4 \\
  \hline
  \end{tabularx}
  \label{Tempolate Result}
\end{table*}

\begin{table*}[t!]
  \normalsize
   \caption{Analysis of different action phase number on ActivityNet v1.3 under the 50\% seen / 50\% unseen split. The phase number adopted in this paper is highlighted with a \colorbox{blue!10}{blue background}.}
  \centering
  \begin{tabularx}{0.9\textwidth}{ >{\raggedright\arraybackslash}p{0.35\linewidth}  >{\centering\arraybackslash}X  >{\centering\arraybackslash}X  >{\centering\arraybackslash}X  >{\centering\arraybackslash}X>{\centering\arraybackslash}p{0.1\linewidth}}
  \hline
   \multirow{2}{*}{\textbf{Phase Number}} & \multicolumn{4}{c}{mAP@tIOU (\%)} & \multirow{2}{*}{Time (min)} \\
   \cmidrule(lr){2-5}
   & 0.3 & 0.5 & 0.7 & Avg & \\
   \hline
    One \textit{(Glob)} & 51.0 & 32.8 & 5.9 & 32.3 & 14.7 \\
    Two \textit{(Start, End)} & 51.9 & 33.7 & 6.6 & 33.1 & 15.1 \\
    Three \textit{(Start, Mid, End)} & 52.6 & 34.5 & 7.1 & 33.9 & 15.8 \\
    \rowcolor{blue!10} Four \textit{(Start, Mid, End, Glob)} & 53.1 & 35.3 & 7.7 & 34.6 & 16.4 \\
    Five \textit{(Start, Mid1, Mid2, End, Glob)} & 53.5 & 35.5 & 8.0 & 34.8 & 17.2\\
    Six \textit{(Start, Mid1, Mid2, Mid3, End, Glob)} & 53.7 & 35.9 & 8.1 & 35.1 & 18.1\\
    \hline
  \end{tabularx}
  \label{Phase Number ACT}
\end{table*}

\subsection{Analysis of Different Phase Number on  ActivityNet v1.3} 
In the main submission, we analyze the impact of different phase number on THUMOS14 dataset. Here, we provide more results on ActivityNet v1.3 dataset. 
As shown in Table~\ref{Phase Number ACT}, performance consistently improves as the number of phases increases, confirming that finer-grained decomposition provides richer semantic cues and benefits unseen action detection. However, the gains become marginal once the phase count exceeds four, reflecting the saturation of informative semantics and the emergence of redundant or noisy descriptions. Moreover, using more phases increases computational cost due to additional alignment operations. Balancing accuracy and efficiency, we therefore adopt a four-phase design (start, middle, end, global) in our final model, which provides an effective trade-off while maintaining strong generalization, consistent with conclusions in the main submission on THUMOS14 dataset.

\begin{table*}[!]
\centering
\caption{Analysis of Different LLM Backbone. The LLM backbone adopted in this paper is highlighted with a \colorbox{blue!10}{blue background}.}
  \normalsize
  \begin{tabularx}{\textwidth}{ 
  >{\centering\arraybackslash}p{0.10\linewidth} 
  >{\centering\arraybackslash}p{0.14\linewidth} 
  >{\centering\arraybackslash}X 
  >{\centering\arraybackslash}X  
  >{\centering\arraybackslash}X 
  >{\centering\arraybackslash}X 
  >{\centering\arraybackslash}X 
  >{\centering\arraybackslash}X  
  >{\centering\arraybackslash}X  
  >{\centering\arraybackslash}X
  }
  \hline
   \multirow{2}{*}{\textbf{Data Split}} & \multirow{2}{*}{\textbf{LLM Backbone}} 
  & \multicolumn{4}{c}{\textbf{THUMOS14}} & \multicolumn{4}{c}{\textbf{ActivityNet v1.3}} \\
  \cmidrule(lr){3-6} \cmidrule(lr){7-10}
  & & 0.3 & 0.5 & 0.7 & Avg & 0.5 & 0.75 & 0.95 & Avg \\
  \hline
 \multirow{4}{*}{\shortstack{50\% Seen \\ 50\% Unseen}} & Qwen3 & 64.9 & 49.1 & 24.0 & 46.2 & 53.3 & 35.2 & 7.5 & 34.3 \\
 & Deepseek v3 & 64.5 & 49.0 & 23.7 & 46.2 & 52.6 & 34.8 & 7.3 & 34.0 \\
  & GPT-4 & 65.1 & 49.4 & 24.1 & 46.6 & 52.9 & 35.3 & 7.6 & 34.5 \\
 &  \cellcolor{blue!10} GPT-4o & \cellcolor{blue!10} 65.4 & \cellcolor{blue!10} 49.7 & \cellcolor{blue!10} 24.3 & \cellcolor{blue!10} 46.9 & \cellcolor{blue!10} 53.1 & \cellcolor{blue!10} 35.3 & \cellcolor{blue!10} 7.7 & \cellcolor{blue!10} 34.6 \\
 \hline
 \multirow{4}{*}{\shortstack{75\% Seen \\ 25\% Unseen}} & Qwen3 & 69.6 & 54.0 & 27.5 & 51.6 & 54.8 & 36.5 & 8.5 & 36.7 \\
 & Deepseek v3 & 69.8 & 53.7 & 27.5 & 51.4 & 55.1 & 36.7 & 8.3 & 37.0 \\
 & GPT-4 & 70.0 & 54.3 & 28.1 & 51.8 & 55.9 & 37.6 & 8.4 & 37.2 \\
  &  \cellcolor{blue!10} GPT-4o & \cellcolor{blue!10} 70.5 & \cellcolor{blue!10} 54.6 & \cellcolor{blue!10} 28.3 & \cellcolor{blue!10} 52.1 & \cellcolor{blue!10} 56.2 & \cellcolor{blue!10} 37.8 & \cellcolor{blue!10} 8.6 & \cellcolor{blue!10} 37.4 \\
  \hline
  \end{tabularx}
  \label{LLM selection}
\end{table*}

\begin{figure*}[!]
\centering
  \includegraphics[width=\textwidth]{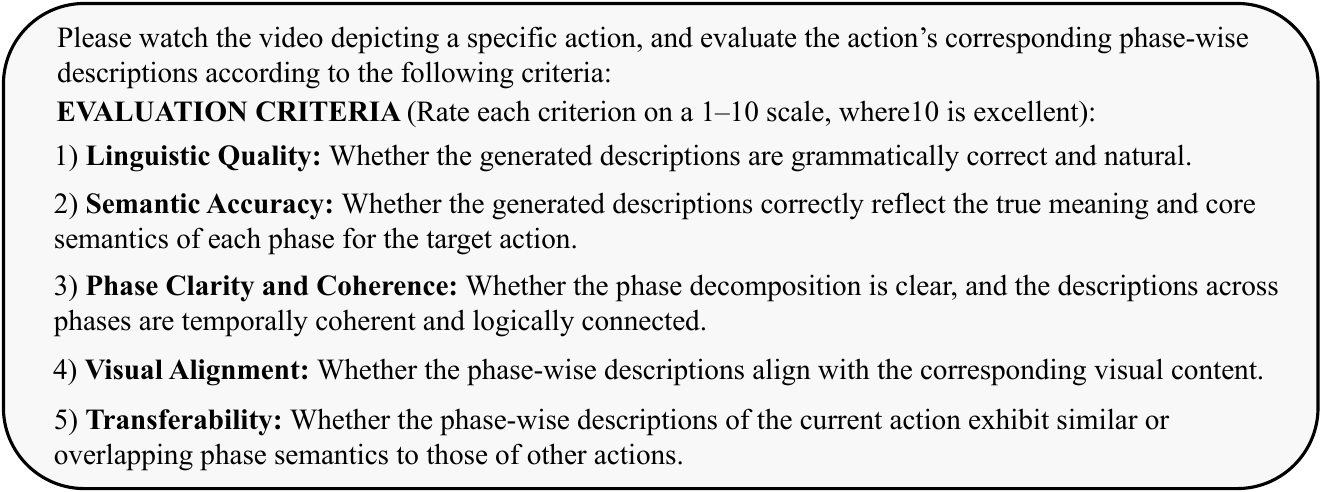}
  \caption{Evaluation Template for GPT-4V and Human Assessments: Rating Across Five Dimensions—Linguistic Quality, Semantic Accuracy, Phase Clarity and Coherence, Visual Alignment, and Transferability.}
  \label{Evaluation Template}
\end{figure*}

\begin{table*}[!]
  \normalsize
  \centering
  \caption{GPT-4V and Human Evaluation of Description Quality on THUMOS14. The best results are highlighted in \textbf{\textcolor{red}{Red}}.}
  \begin{tabularx}{0.9\textwidth}{
  >{\raggedright\arraybackslash}p{0.09\linewidth}
  >{\centering\arraybackslash}p{0.13\linewidth}
  >{\centering\arraybackslash}X
  >{\centering\arraybackslash}X
  >{\centering\arraybackslash}X
  >{\centering\arraybackslash}X
  >{\centering\arraybackslash}X}
  \hline
   \multirow{2}{*}{Evaluation} & \multirow{2}{*}{Method} & \multicolumn{5}{c}{Rating [score/10.00] ($\uparrow$)} \\
   \cmidrule(lr){3-7}
   & & Quality & Accuracy & Coherence & Alignment & Transferability \\
  \hline
  \multirow{2}{*}{GPT-4V} & Ground Truth 
  & 9.03 
  & \textcolor{red}{\textbf{8.35}} 
  & 8.43  
  & \textcolor{red}{\textbf{8.15}} 
  & 7.97 \\
  
  & \textbf{Ours} 
  & \textcolor{red}{\textbf{9.21}} 
  & 8.22 
  & \textcolor{red}{\textbf{8.72}} 
  & 7.86 
  & \textcolor{red}{\textbf{8.49}} \\
  
  \hline
  
  \multirow{2}{*}{Human} & Ground Truth 
  & \textcolor{red}{\textbf{8.77}} \textcolor{red}{$\pm 0.09$}
  & \textcolor{red}{\textbf{8.10}} \textcolor{red}{$\pm 0.15$}
  & 8.23 $\pm 0.10$
  & \textcolor{red}{\textbf{7.91}} \textcolor{red}{$\pm 0.07$}
  & 7.71 $\pm 0.06$  \\
  
  & \textbf{Ours} 
  & 8.46 $\pm 0.06$
  & 7.95 $\pm 0.08$
  & \textcolor{red}{\textbf{8.39}} \textcolor{red}{$\pm 0.12$}
  & 7.64 $\pm 0.09$
  & \textcolor{red}{\textbf{8.28}} \textcolor{red}{$\pm 0.10$} \\
  
  \hline
  \end{tabularx}
  \label{Evaluation}
\end{table*}

\subsection{Analysis of Different LLM Backbone}
We further extend the analysis of different LLM backbones beyond the THUMOS14 dataset under the 50\% seen / 50\% unseen split in the main submission. Specifically, we additionally evaluate ActivityNet v1.3 dataset under both the 50\% seen / 50\% unseen and 75\% seen / 25\% unseen splits, as well as THUMOS14 under the 75\% seen / 25\% unseen split. The results on four widely used LLMs: Qwen3~\cite{yang2025qwen3}, Deepseek v3~\cite{liu2024deepseek}, GPT-4~\cite{achiam2023gpt}, and GPT-4o~\cite{hurst2024gpt} are shown in Table~\ref{LLM selection}. we observe that overall performance across both datasets and evaluation splits remains relatively stable, regardless of the LLM backbone used. The relatively minor differences among the LLMs further suggest that our method is robust to the choice of LLM backbone, which is desirable for practical deployment.

\begin{figure*}[t]
\centering
  \includegraphics[width=\textwidth]{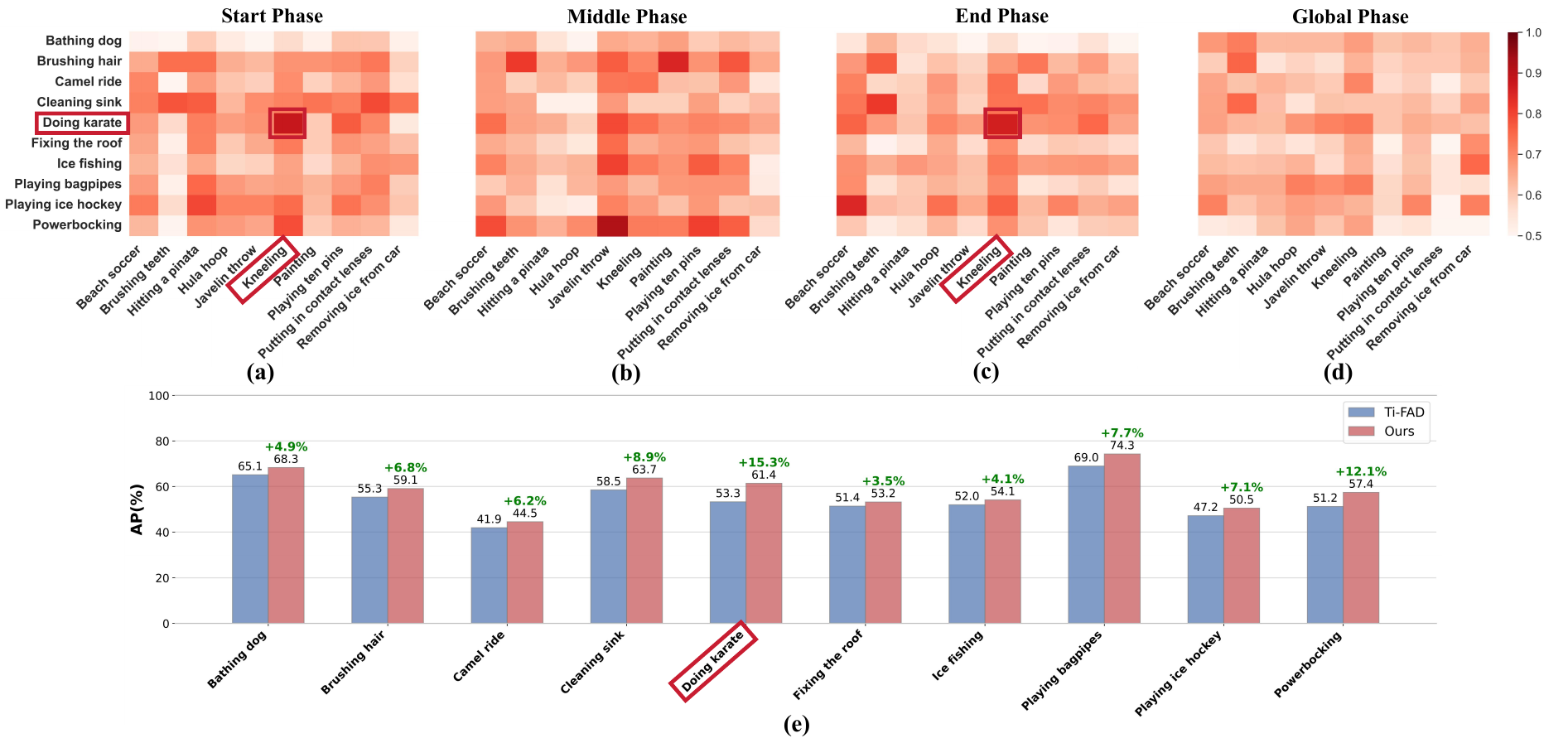}
  \caption{Phase-wise Semantic Similarity ((a)-(d)) and Per-unseen class AP (\%) (e) at tIoU threshold 0.5 on  ActivityNet v1.3 under the 50\% seen / 50\% unseen split. For (a)-(d), the vertical denotes unseen (testing) classes, the horizontal denotes seen (training) classes.}
  \label{AP}
\end{figure*}

\begin{table*}[!]
  \normalsize
  \caption{Plug-and-Play Capability of the Proposed PDA on THUMOS14  under the 50\% seen / 50\% unseen split.}
  \centering
  \begin{tabularx}{0.8\textwidth}{ >{\raggedright\arraybackslash}p{0.18\linewidth}  >{\centering\arraybackslash}p{0.12\linewidth}  >{\centering\arraybackslash}X  >{\centering\arraybackslash}X  >{\centering\arraybackslash}X  >{\centering\arraybackslash}X}
  \hline
   \multirow{2}{*}{Backbone} & \multirow{2}{*}{Method} & \multicolumn{4}{c}{mAP@tIOU (\%)} \\
   \cmidrule(lr){3-6}
   & & 0.3 & 0.5 & 0.7 & Avg \\
  \hline
  \multirow{2}{*}{DyFADet (ECCV'24)} & Baseline & 17.5 & 12.2 & 5.7 & 11.9 \\
  & \textbf{PDA} & \textcolor{red}{\textbf{66.1}} & \textcolor{red}{\textbf{50.2}} & \textcolor{red}{\textbf{25.0}} & \textcolor{red}{\textbf{47.5}} \\
  \hline
  \multirow{2}{*}{DiGIT (CVPR'25)} & Baseline & 19.1 & 13.5 & 6.1  &  13.0 \\
  & \textbf{PDA} & \textcolor{red}{\textbf{67.3}} & \textcolor{red}{\textbf{50.7}} & \textcolor{red}{\textbf{25.2}} & \textcolor{red}{\textbf{48.0}} \\
  
    \hline
  \end{tabularx}
  \label{Versatility}
\end{table*}

\begin{table}[t!]
  \normalsize
   \caption{Effectiveness of varying visual/textual backbones on THUMOS14.}
  \begin{tabularx}{\columnwidth}{ >{\raggedright\arraybackslash}X  >{\centering\arraybackslash}p{0.14\linewidth}  >
  {\centering\arraybackslash}p{0.14\linewidth}  >{\centering\arraybackslash}p{0.18\linewidth}  >{\centering\arraybackslash}p{0.18\linewidth}}
  \hline
   \multirow{2}{*}{Method} & \multicolumn{2}{c}{Feature} & \multicolumn{2}{c}{mAP@AVG} \\
   \cmidrule(lr){2-3} \cmidrule(lr){4-5}
   & Visual & Text & 50\%-50\% & 75\%-25\% \\
   \hline
   \multirow{4}{*}{Ti-FAD} & CLIP-B & CLIP-B & 27.3 & 29.7 \\
   & CLIP-L & CLIP-L & 27.2 & 30.6 \\
   & I3D & CLIP-B & 41.2 & 46.8 \\
   & I3D & CLIP-L & 40.6 & 47.3 \\
   \hline
   \multirow{4}{*}{Ours} & CLIP-B & CLIP-B & \textcolor{red}{\textbf{30.1}} & \textcolor{red}{\textbf{33.5}} \\
   & CLIP-L & CLIP-L & \textcolor{red}{\textbf{30.6}} & \textcolor{red}{\textbf{34.3}} \\
   & I3D & CLIP-B & \textcolor{red}{\textbf{46.9}} & \textcolor{red}{\textbf{52.1}} \\
   & I3D & CLIP-L & \textcolor{red}{\textbf{47.2}} & \textcolor{red}{\textbf{52.9}}\\
  \hline
  \end{tabularx}
  \label{backbone}
\end{table}

\begin{table*}[!]
\centering
\caption{Comparison with Previous LLM-base Label Expansion Methods.}
  \normalsize
  \begin{tabularx}{\textwidth}{ >{\arraybackslash}p{0.22\linewidth}  >
  {\arraybackslash}p{0.46\linewidth}  >
  {\centering\arraybackslash}X 
  }
  \hline
 Task & Characteristics & Summary \\
 \hline
 \multirow{3}{*}{Action Localization~\cite{aklilu2024zero, zheng2024training}} & Decompose actions into defining attributes and aggregate these attributes to align with frame-level embeddings, enabling more precise localization. & \textit{\textbf{\multirow{6}{*}{\shortstack{Enrich textual semantics for \\ more concise alignment.}}}}\\
  \cmidrule(lr){1-2}
 \multirow{3}{*}{Action Recognition~\cite{bosetti2024text, jia2024generating}} & Decompose actions into multi-dimensional descriptions and aggregate these descriptions to align with averaged visual embeddings for more precise recognition.& \\
 \hline
 \multirow{4}{*}{\textbf{Ours}} & Decompose actions into \textit{\textbf{multi-phase descriptions}} and adaptively perform \textit{\textbf{phase-wise alignment}} with visual features to learn transferable action patterns, enhancing zero-shot action detection performance. & \textit{\textbf{\multirow{4}{*}{\shortstack{Learn transferable action \\ knowledge for generalized \\ zero-shot detection.}}}}\\
  \hline
\end{tabularx}
  \label{LLM label expansion summary}
\end{table*}

\begin{table*}[!]
\centering
\caption{Comparison with  LLM-base Label Expansion under the 50\% seen / 50\% unseen split.}
  \normalsize
  \begin{tabularx}{\textwidth}{ >{\centering\arraybackslash}p{0.10\linewidth}  >
  {\centering\arraybackslash}p{0.16\linewidth}  >
  {\centering\arraybackslash}X  >
  {\centering\arraybackslash}X  >
  {\centering\arraybackslash}X  >
  {\centering\arraybackslash}X  >
  {\centering\arraybackslash}X  >
  {\centering\arraybackslash}X  >
  {\centering\arraybackslash}X  >
  {\centering\arraybackslash}X
  }
  \hline
   \multirow{2}{*}{\textbf{Expansion}} & \multirow{2}{*}{\textbf{Method}} 
  & \multicolumn{4}{c}{\textbf{THUMOS14}} & \multicolumn{4}{c}{\textbf{ActivityNet v1.3}} \\
  \cmidrule(lr){3-6} \cmidrule(lr){7-10}
  & & 0.3 & 0.5 & 0.7 & Avg & 0.5 & 0.75 & 0.95 & Avg \\
  \hline
  Baseline & - & 56.2 & 42.7 & 20.4 & 40.3 & 49.7 & 31.5 & 4.9 & 31.2 \\
  \hline
 \multirow{2}{*}{\shortstack{Global Label \\ Expansion}} & (a) w/o Decompose & 60.1 & 46.5 & 22.2 & 43.4 & 51.7 & 33.2 & 6.3 & 32.7 \\
 & (b) w/ Decompose & 61.1 & 47.3 & 22.7 & 44.0 & 51.8 & 33.5 & 6.6 & 32.9\\
 \hline
 \multirow{4}{*}{\shortstack{Single-phase \\ Expansion}} & Start & 57.4 & 43.8 & 21.1 & 41.2 & 50.1 & 32.0 & 5.2 & 31.6 \\
 & Middle & 58.1 & 44.2 & 21.4 & 41.5 & 50.8 & 32.4 & 5.7 & 32.0 \\
 & End & 57.8 & 44.6 & 21.5 & 41.7 & 50.6 & 32.5 & 5.3 & 31.8 \\
 & Global & 59.3 & 45.5 & 21.9 & 42.5 & 51.0 & 32.8 & 5.9 & 32.3\\
  \hline
\textbf{Ours} & - & \textcolor{red}{\textbf{65.4}} & \textcolor{red}{\textbf{49.7}} & \textcolor{red}{\textbf{24.3}} & \textcolor{red}{\textbf{46.9}} & \textcolor{red}{\textbf{53.1}} & \textcolor{red}{\textbf{35.3}} & \textcolor{red}{\textbf{7.7}} & \textcolor{red}{\textbf{34.6}} \\
\hline
  \end{tabularx}
  \label{LLM label expansion}
\end{table*}

\subsection{Analysis of Description Quality}
In this paper, we leverage the chain-of-thought (CoT) capability of LLMs to decompose action labels into coherent multi-phase descriptions. To evaluate the quality of these generated descriptions, we conduct both GPT-based and human-based subjective assessments. For GPT evaluation, we employ the multimodal LLM GPT-4V~\cite{yang2023dawn}, and for human evaluation, we recruit ten volunteers.
First, several domain experts are invited to decompose each action label into start, middle, end, and global descriptions, which serve as the ground truth.
Subsequently, GPT-4V and human volunteers are employed to evaluate the generated descriptions across five dimensions: Linguistic Quality, Semantic Accuracy, Phase Clarity and Coherence, Visual Alignment, and Transferability.
The detailed evaluation protocol is illustrated in Figure~\ref{Evaluation Template}. For each action label, two corresponding videos are randomly selected for assessment. To alleviate potential scoring bias among human evaluators, we further compute the confidence levels of their ratings. The aggregated results are reported in Table~\ref{Evaluation}. 
Experimental results show that the generated descriptions achieve comparable performance to human-decomposed ones in terms of linguistic quality, semantic accuracy, phase clarity, and visual alignment, indicating that the generated descriptions are linguistically natural and effectively capture the underlying action semantics. Notably, the generated descriptions score higher on transferability metric, indicating a stronger capacity to capture cross-action phase regularities learned from large-scale textual knowledge, which could further enhance zero-shot generalization in open-vocabulary temporal action detection.

\section{More Experimental Results}

\subsection{Plug-and-Play Capability of the Proposed PDA}
To assess the generalization and versatility of the proposed Phase-wise Decomposition and Alignment (PDA) framework, we conduct plug-and-play experiments on two recent closed-set TAD models, DyFADet~\cite{yang2024dyfadet} and DiGIT~\cite{kim2025digit}. We integrate the PDA modules into these models and evaluate whether the introduced decomposition and alignment mechanisms improve their performance under the open-vocabulary setting. As reported in Table~\ref{Versatility}, our method consistently surpasses the baselines directly adapted to the OV-TAD protocol in~\cite{li2024detal,wangconcept}, achieving substantial gains in recognizing unseen action categories. These results demonstrate that PDA—through LLM-based multi-phase semantic decomposition followed by adaptive phase-wise alignment—effectively learns transferable action patterns and could serve as a versatile, plug-and-play component that enhances the generalization capability of existing closed-set TAD models in open-vocabulary scenarios.

\subsection{Effects of Different Visual/Textual Backbones} 
In this section, we investigate the robustness of our approach under different visual and textual backbones.
For the visual encoder, following~\cite{lee2024text}, we consider CLIP ViT-B/16, ViT-L/14 and I3D. 
For the textual encoder, we consider CLIP ViT-B/16 and ViT-L/14. 
We compare against the global-alignment SOTA method Ti-FAD, which directly matches label-level semantics with global visual representations.
As shown in Table~\ref{backbone}, our method consistently surpasses Ti-FAD across all combinations of visual and textual encoders.
This demonstrates the universality of our approach and its capacity to generalize across diverse backbone settings, thereby supporting improved open-vocabulary temporal action detection.

\begin{figure*}[t]
\centering
  \includegraphics[width=\textwidth]{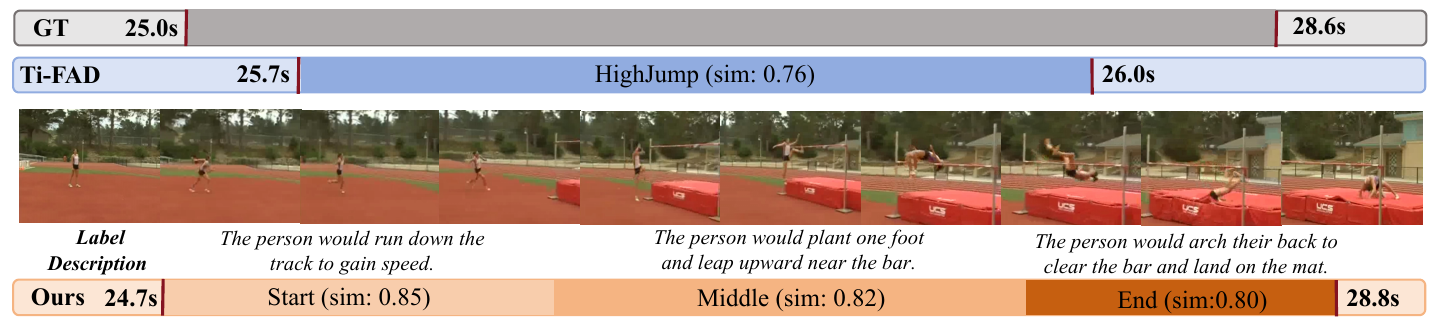}
  \caption{Visualization of the detection results “HighJump” on THUMOS14 under the 50\% seen / 50\% unseen split. “sim” represents the visual-textual similarity at each phase.}
  \label{HighJump}
\end{figure*}

\begin{figure*}[t]
\centering
  \includegraphics[width=\textwidth]{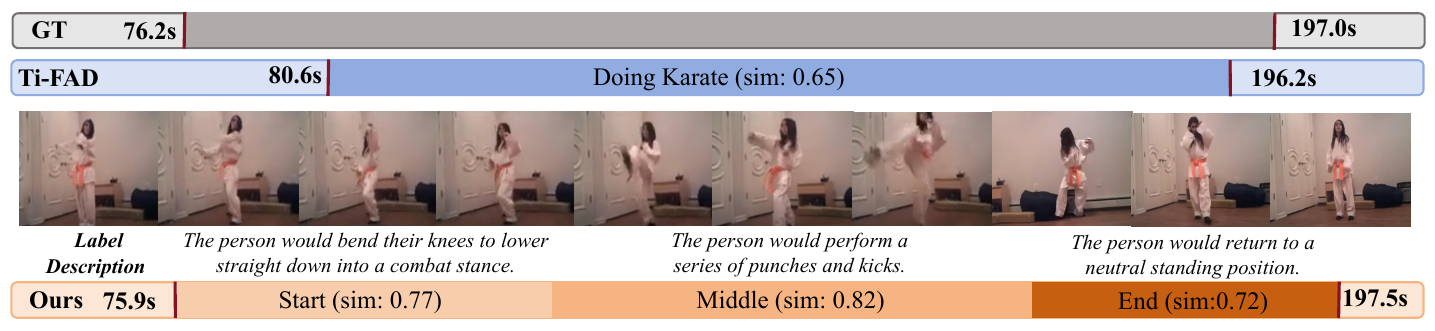}
  \caption{Visualization of the detection results “Doing Karate” on ActivityNet v1.3 under the 50\% seen / 50\% unseen split. “sim” represents the visual-textual similarity at each phase.}
  \label{Doing Karate}
\end{figure*}

\begin{figure*}[!]
\centering
  \includegraphics[width=\textwidth]{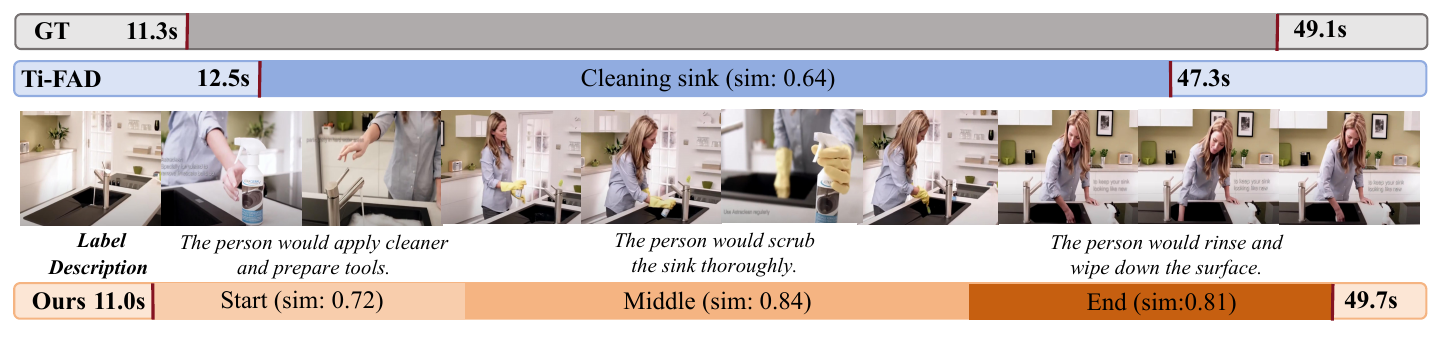}
  \caption{Visualization of the detection results “Cleaning sink” on ActivityNet v1.3 under the 50\% seen / 50\% unseen split. “sim” represents the visual-textual similarity at each phase.}
  \label{Cleaning sink}
\end{figure*}

\subsection{Difference Against Previous LLM-base Label Expansion Methods}
As summarized in Table~\ref{LLM label expansion summary}, some previous methods~\cite{ju2023multi,aklilu2024zero,zheng2024training,bosetti2024text} also use LLM to expand labels, but their goal is to generate more detailed semantic descriptions for visual matching. However, our approach leverages LLMs to extract transferable knowledge across semantically diverse labels. Combined with adaptive phase-wise alignment, this enables the discovery of phase-level transferable action patterns, thereby enhancing zero-shot detection.
To clarify the source of our performance gain and distinguish our method from simple LLM-based label expansion, we design two comparison variants: 1) Global Label Expansion includes two sub-variants: (a) using GPT-4o to generate detailed action descriptions without phase decomposition; (b) generating start, middle, end, and global phase descriptions with GPT-4o and concatenating them into a single expanded label. These enriched textual descriptions are aligned with video features using a global alignment strategy, similar to prior LLM-augmented methods. 2) Single-Phase Expansion leverages phase-specific semantic descriptions (Start, Middle, End, Global) to match video features individually. 
As shown in Table~\ref{LLM label expansion}, global label expansion yields only marginal improvements, indicating that \textbf{richer semantics alone offer limited gains}. The single-phase variant performs even worse. In contrast, our full PDA framework achieves significant improvements, confirming that \textbf{the core advantage stems} not from single LLM-based label expansion but \textbf{from the combination with adaptive phase-wise visual–textual alignment}, which effectively transfers fine-grained visual priors from seen to unseen actions.

\subsection{Phase-wise Semantic Similarity on Activitynet v1.3.}
We further evaluate the effectiveness of PDA on ActivityNet v1.3 by analyzing phase-level semantic similarity and reporting per-class AP for 10 randomly selected seen and unseen classes.
Figures~\ref{AP} (a)-(d) present phase-wise semantic similarity matrices between seen and unseen classes, darker red regions indicate higher cosine similarity between corresponding phase descriptions. Notably, several unseen actions exhibit strong semantic alignment with seen actions at specific phases, despite differing at the global action level.
A representative example is the seen-unseen pair Kneeling and Doing karate, which display high semantic similarity in both the start and end phases. Both actions begin with a similar preparatory motion—\textit{“The person would bend their knees to lower straight down toward the ground”} (Kneeling) versus \textit{“The person would bend their knees to lower straight down into a combat stance”} (Doing karate). Their ending phases also involve returning to an upright position. Such shared phase patterns provide transferable cues that our model effectively leverages during inference on unseen classes.
In addition, Figure~\ref{AP} (e) reports per-class Average Precision (AP) comparisons between our method and Ti-FAD. Consistent with results on THUMOS14, our approach achieves higher AP across all unseen categories, particularly those with strong phase-level semantic affinity to the seen set (e.g., Doing karate). These findings further validate that adaptive phase-aware decomposition promotes more effective knowledge transfer and enhances generalization to previously unseen actions.

\subsection{More Qualitative Results}
To further demonstrate the effectiveness of our proposed framework, we present additional qualitative results on one unseen action class (HighJump) from THUMOS14 and two unseen classes (Doing karate and Cleaning sink) from ActivityNet v1.3. As shown in Figure~\ref{HighJump},\ref{Doing Karate},\ref{Cleaning sink}, our method consistently yields more accurate temporal boundaries and higher phase-level matching scores compared to the baseline Ti-FAD. Notably,  the semantics of the corresponding phase-specific textual descriptions exhibit strong alignment with the predicted segments , indicating that our model not only improves classification accuracy but also enhances localization precision across datasets and action types. These consistent improvements further underscore the generalizability of our approach in recognizing unseen categories through fine-grained visual-semantic reasoning.
